\documentclass[nohyperref]{article}

\usepackage{microtype}
\usepackage{graphicx}
\usepackage{booktabs} 

\usepackage{hyperref}

\usepackage[accepted]{icml2022}

\usepackage{amsmath}
\usepackage{amssymb}
\usepackage{mathtools}
\usepackage{amsthm}
\usepackage{subcaption}
\usepackage{bbm}
\usepackage{multicol}
\usepackage{spverbatim}
\usepackage{fancyvrb}
\usepackage{fvextra}

\usepackage[capitalize,noabbrev]{cleveref}

\theoremstyle{plain}

\theoremstyle{definition}

\theoremstyle{remark}

\newcounter{numquote}

\usepackage[textsize=tiny]{todonotes}

\icmltitlerunning{Policy Optimization with Sparse Global Contrastive Explanations}

\begin{document}

\twocolumn[
\icmltitle{Policy Optimization with Sparse Global Contrastive Explanations}



\icmlsetsymbol{equal}{*}

\begin{icmlauthorlist}
\icmlauthor{Jiayu Yao}{harvard}
\icmlauthor{Sonali Parbhoo}{harvard,imp}
\icmlauthor{Weiwei Pan}{harvard}
\icmlauthor{Finale Doshi-Velez}{harvard}

\end{icmlauthorlist}

\icmlaffiliation{harvard}{SEAS, Harvard University}
\icmlaffiliation{imp}{Imperial College London}

\icmlcorrespondingauthor{Jiayu Yao}{jiy328@g.harvard.edu}

\icmlkeywords{Machine Learning, ICML}

\vskip 0.3in
]



\printAffiliationsAndNotice{}  

\begin{abstract}
  We develop a Reinforcement Learning (RL) framework for improving an existing behavior policy via sparse, user-interpretable changes.  Our goal is to make minimal changes while gaining as much benefit as possible.  We define a minimal change as having a sparse, global contrastive explanation between the original and proposed policy.  We improve the current policy with the constraint of keeping that global contrastive explanation short. We demonstrate our framework with a discrete MDP and a continuous 2D navigation domain. 
\end{abstract}

\section{Introduction}\label{sec:intro}
Human-understandable descriptions of an RL agent's policy are important for human oversight, especially in safety-critical settings~\citep{rudin2021interpretable}. Contrastive explanations are known to be effective in helping people understand machine learning models~\citep{jacovi2021contrastive,rathi2019generating,vanderwaa2018contrastive,van2018contrastive}. In the context of RL, contrastive explanations describe the differences in outcomes between the current policy and an alternate one~\citep{puiutta2020explainable}.  By leveraging the context of the current policy, the outputs of contrastive explanations are often sparser and thus less cognitively demanding than a complete explanation of the new policy~\citep{miller2019explanation, du2019techniques}. When the new policy is very different from the current one, the contrastive explanations can still be dense. 
Existing works show that in this case, we still want contrastive explanations to be sparse, which are more understandable and executable  for humans~\cite{du2019techniques}. 

In this work, we build a framework that finds an optimal set of improvements to the current policy (with which the user is already familiar) such that the contrastive explanation given the current and improved policy remains sparse. Our framework consists of two parts: (1) an explanation-generator that creates global contrastive explanations for two RL policies---to our knowledge, this is the first comprehensive contrastive explanation system for RL---and (2) an optimization procedure for improving the current policy while explicitly constraining the length of the associated explanation (given by the explanation-generator). 

\textbf{Framework Part I: Explanation Generation } We consider contrastive explanations of the form: 
\begin{flalign}
\small
\begin{tabular}{@{}p{0.42\textwidth}}
\label{explanation_form}
``Given the current state, doing action $A_1$ instead of 
$B_1$ when you reach state $1$, 
then doing action $A_2$ instead of $B_2$ when you reach state $2$, $\dots$, then finally doing
$A_m$ instead of $B_m$ when you reach state $m$, will lead 
to better / worse / unknown differences in outcomes."
\end{tabular}
\end{flalign}
For example, imagine a mobile fitness app that helps users build workout routines. We might explain a new workout plan with ``If you mostly do cardio [current state], then after cardio [state $1$], doing weightlifting [$A_1$] instead of stretching [$B_1$] will improve fat loss [outcome difference]"---this can be more effective than simply describing the new policy, especially if the new policy overlaps significantly with the old policy.  While other works
have considered contrastive explanations in RL~\citep{vanderwaa2018contrastive,sukkerd2020tradeoff}, to our knowledge, ours is the first to lay out a \emph{complete} contrast definition that includes the \emph{series} of relevant states and decisions as well as expected changes to future outcomes.

\textbf{Contribution 1: A method to generate usable contrastive explanations, taking into account scalability, the space of relevant outcomes, and off-policy outcome estimation.}

\emph{Scalability.} 
Large, continuous domains are common in real life. For example, a fitness app may collect users' demographical information, vital signs, sleeping schedules, etc. This creates two challenges for generating global contrastive explanations: (1) it is computationally expensive to compare policies on all states; (2) it is impractical to explain each and every state to users. 
\emph{To scale the explanation-generator to large, continuous domains, we develop an approach to perform efficient state exploration and appropriate state abstractions} (Section~\ref{subsec:state_identifier} \&~\ref{subsec:state_aggregator}). 

\emph{Relevant outcomes.} 
Current interpretable RL works tend to focus on explaining the optimized outcome (such as the reward function or the Q-value) \citep{verma2018programmatically,liu2018toward,DBLP:journals/corr/abs-1903-09338,DBLP:journals/corr/abs-2011-05064}. But in the real world, humans care about multiple outcomes, some of which are often not explicitly considered in the optimization and thus not explained~\citep{doshi2017towards,dovsilovic2018explainable}. For example, an algorithm may aim to increase users' physical activity overall, while users may have more specific goals such as reducing body fat and improving cardiovascular health. \emph{To make the explanations more relevant to users, we allow users to specify a customized set of outcomes} (Section~\ref{subsec:outcome_estimator}).

\emph{Off-policy outcome estimation.}
In the setting where only batch data is given, we must be sensitive to two considerations. First, without further interaction with the environment, it is hard to compute the outcomes of the contrasting policy. For example, when proposing a new workout routine, we only have access to the user's past data. Until the new routine is tried out, we will not know the actual fat reduction and cardiovascular health improvement rates. Furthermore, in cases where available data is scarce, it is important to communicate that the model is unsure of the outcomes~\citep{kompa2021second}. \emph{In batch settings, we use off-policy evaluation ~\citep{hanna2017bootstrapping} to estimate the outcome of the contrasting policy and report confidence intervals.} (Section~\ref{subsec:outcome_estimator}).

\begin{figure}
    \centering
    \includegraphics[scale=0.25]{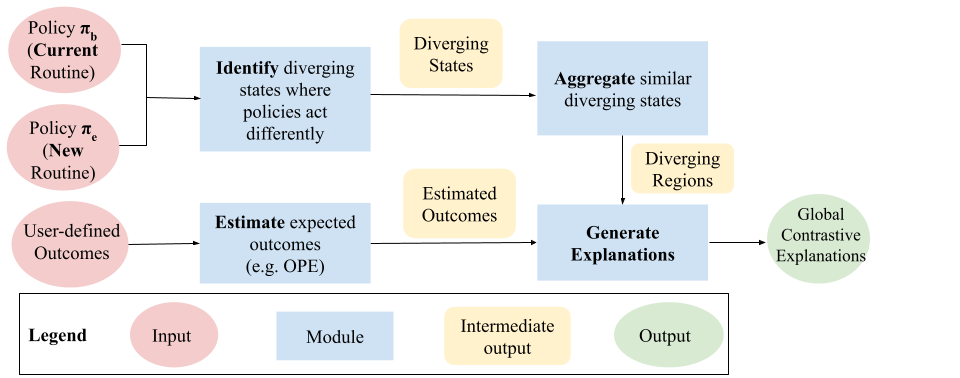}
    \caption{Our explanation-generator takes in two policies, compares two policies in terms of their decisions and expected outcomes, and finally creates a global contrastive explanation by combining the behavioral and the outcome differences. }
    \label{fig:framework}
\end{figure}

\emph{Complete Explanation Generator.}  We build an explanation-generator that includes all the above key features. The complete procedure is illustrated in Figure~\ref{fig:framework}. 
The generator takes in two policies, either in the batch form or the functional form. It combines the appropriate state abstractions and the estimated outcomes to generate a global contrastive explanation of two given policies (Section~\ref{subsec:explanation_generator}).

\textbf{Framework Part II: Explanation-Sparsity-Constrained Optimization } The explanation-generator in Part I provides us with a way to compactly explain a new policy in terms of changes to the current one. However, if the two policies differ greatly, the explanation in Form \ref{explanation_form} will be long. In real life, users often prefer sparse explanations because: (1) sparsity lessens cognitive demand; (2) users prefer making minimal changes to their current routine while maximizing benefit gain. Thus, when we solve for an optimal new policy, we want to constrain our choices to policies that generate sparse contrastive explanations given our current policy.

\textbf{Contribution 2: An optimization procedure for improving the current policy that keeps contrastive explanations sparse. } We recast policy optimization as the problem of learning how to modify the current policy: we associate a cost to each change to the current policy and find a set of changes that optimizes a single objective, while keeping the total cost within a fixed budget (thus capping the complexity of the associated contrastive explanation). We formalize this optimization as a Constrained Markov Decision Process problem. Note that here, the objective being optimized may or may not be one of the user-defined outcomes in Framework Part I: Explanation Generation. 

\textbf{Empirical Results }We demonstrate the efficacy of our framework on two domains: a discrete toy domain, a continuous 2D navigation domain. 
We show that for explanation generation, our approach generates explanations that are more concise, complete and interpretable comparing to other contrastive explanation forms; for optimization, our optimization converges to policies with sparser contrastive explanations compared to naive policy optimization methods such as policy iteration.

\section{Related Work}\label{sec:related_work}

\textbf{Interpretable RL.} There is a large body of works focusing on learning interpretable policies and explaining the agents' action at a specific decision point \citep{verma2018programmatically,liu2018toward,DBLP:journals/corr/abs-1903-09338,DBLP:journals/corr/abs-2011-05064}. We consider the task of explaining the difference between \emph{two} policies instead of explaining each policy individually.

We also note that contrastive explanations are different from counterfactual explanations. Counterfactual explanations ask how to alter the past to realize the changes in the outcome, while contrastive explanations ask why one outcome occurs instead of the other~\cite{mcgill1993contrastive}.

\textbf{Contrastive Explanations for non-RL tasks.} Several recent works focus on generating contrastive explanations for a specific input (local explanations) in single-step decision-making tasks\citep{vanderwaa2018contrastive,jacovi2021contrastive}. 
Unlike these methods, we focus on sequential decision-making tasks where \emph{multiple} differences might be encountered in one trajectory. We aim to generate a \emph{global} summary of how \emph{two} policies differ across the entire state space. 

\textbf{Contrastive Explanations for RL tasks.} A number of works have explored contrastive explanations for RL. For example, in \citeauthor{vanderwaa2018contrastive}, explanations enumerate all states that appear in the trajectory generated by one policy but not in the trajectory generated by the other. In contrast, we only focus on states where the two policies behave differently, for more concise explanations. 
Furthermore, \citeauthor{vanderwaa2018contrastive} focus on immediate outcomes while we focus on long-term outcomes. 
As another example, \citeauthor{sukkerd2020tradeoff} study a multi-objective setting and help users understand the specific trade-offs between competing objectives of two policies. However, they don't explain \textit{why} objectives are different.
In contrast, our explanations are both more complete and compact.

\textbf{Sparsity in RL Optimization.} 
There is a body of works focusing on generating sparse counterfactual explanations~\cite{wachter2017counterfactual,karimi2020model,spangher2018actionable} by either minimizing the distance between the current instance and the counterfactual instance or associating costs to changes of actions.
To our knowledge, we are the first to consider performing optimization with respect to the contrastive explanations in an RL setting. That is, we optimize a policy to be minimally different from an existing policy. Similar to prior works, we also associate penalties to changes to the current policy.

\begin{figure*}
     \centering
     \begin{subfigure}[b]{0.3\textwidth}
         \centering
         \includegraphics[width=\textwidth]{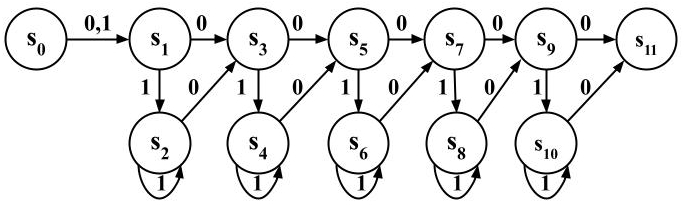}
         \caption{Toy Domain MDP}
         \label{fig:toy_mdp}
     \end{subfigure}
     \hfill
     \begin{subfigure}[b]{0.3\textwidth}
         \centering
         \includegraphics[width=\textwidth]{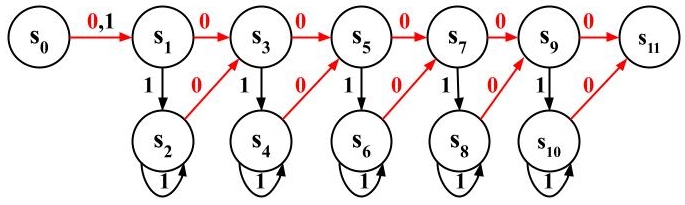}
         \caption{Current Policy $\pi_b$}
         \label{fig:pi_b}
     \end{subfigure}
     \hfill
     \begin{subfigure}[b]{0.3\textwidth}
         \centering
         \includegraphics[width=\textwidth]{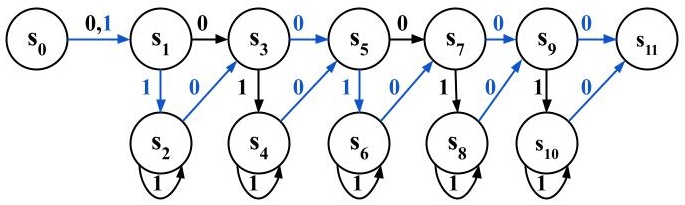}
         \caption{New $\pi_e$}
         \label{fig:pi_e}
     \end{subfigure}
        \caption{Discrete MDP: (a) Definition of the discrete MDP (b) Current Policy $\pi_b$ with actions highlighted by red arrows (c) New Policy $\pi_e$ to which we want to compare with actions highlighted by blue arrows.}
        \label{fig:toy_domain}
\end{figure*}

\section{Notation \& Background}\label{sec:notation}

\textbf{Markov Decision Processes.} 
In RL, an agent learns to maximize a reward signal through interactions with the environment~\citep{sutton2018reinforcement}. An RL problem can be formalized as a Markov Decision Process (MDP) -- defined as a tuple $(\mathcal{S},\mathcal{A},\mathcal{T},\mathcal{R})$, where $\mathcal{S}$ is the state space, $\mathcal{A}$ is the action space that we assume to be discrete, $\mathcal{T}(s'|s,a)$ is the transition dynamics denoting the probability of reaching state $s'$ by taking action $a$ in state $s$, and $\mathcal{R}(s,a)$ is the reward function that returns the immediate reward of performing action $a$ in state $s$.
We assume the initial state $s_0$ is drawn from a distribution $p_0(s)$ and there exists a set of absorbing states $s_T\in\mathcal{S}_T$ with $\mathcal{R}(s_T,\cdot)=0$. A policy $\pi(a|s)$ defines a distribution over actions given state $s$. 

During optimization, we maximize a single objective -- the expected undiscounted return, defined as:
\begin{equation}\label{eqn:expected_return}
   \mathcal{J}_\pi =\mathbb{E}_{\pi}\left[\displaystyle\sum_{t=0}^\infty \mathcal{R}(s_t,a_t)\right].
\end{equation}

\textbf{User Defined Outcomes.} 
Users may also want to know how a policy affects other outcomes.  For example, while a clinician may be mainly interested in developing a policy for treating hypotension, they may also be interested in how that policy affects the patient's respiration, mobility, etc.

We denote the user-defined outcomes as a set consisting of $M$ functions of the current state-action pair, $\textbf{g} = [g^{(0)}(s, a),\dots\ ,g^{(M-1)}(s, a)]^\intercal$ with $g^{(m)}(s_T, \cdot)=0$. The expected outcomes for policy $\pi$ is denoted as $\mathbf{G}_{\pi} = [G^{(0)}_{\pi},\dots,G^{(M-1)}_{\pi}]^\intercal$ where
\begin{equation}\label{eqn:expected_outcomes}
\small
    G^{(m)}_{\pi} = \mathbb{E}_{\pi}\left[\sum_{t=0}^{\infty} g^{(m)}(s_t,a_t)\right]\quad \forall m\in\{0,\dots,M-1\}
\end{equation}

We emphasize that the expected outcomes $\mathbf{G}_\pi$ may be independent of the objective $\mathcal{J}_\pi$ that the RL agent is optimizing. 

In an online RL setting, both the expected return $\mathcal{J}_\pi$ and the expected outcomes $\mathbf{G}_\pi$ can be computed by rolling out trajectories following the policy $\pi$. In a batch setting, where we only have access to a batch of data $\mathcal{D}_b$ collected with the current policy $\pi_b$, both terms can be estimated using off-policy evaluation.
We denote the estimator of the expected return as $\hat{\mathcal{J}_\pi}$ and the estimator of the expected outcomes as $\hat{\mathbf{G}}_\pi$. 
We assume that the biases of the estimators $\hat{\mathbf{G}}_{\pi}$, $\hat{\mathcal{J}}_{\pi}$ are sufficiently small so that they do not affect the quality of the contrastive explanations and the optimization performance.

\textbf{Constrained MDPs.}
\label{subsec:cmdp_wt_l}
In this work, we study planning under interpretability constraints. That is, we seek optimal changes to our current policy that correspond to sparse contrastive explanations.  In Section~\ref{sec:opt}, we will formalize this process as planning under a \emph{constrained MDP} (CMDP).

In an unconstrained MDP setting, the goal is to find a policy $\pi^*$ that maximizes the expected return $\mathcal{J}_\pi$ in Equation~\ref{eqn:expected_return}:
\begin{equation*}
 \pi^* = \arg\max_{\pi} \mathcal{J}_\pi
\end{equation*}
In addition to the standard MDP tuple, a CMDP contains a cost function $ \mathcal{C}(s, a)$ with $\mathcal{C}(s_T,\cdot)=0$ and a threshold $\kappa$ (in our context, the cost $\mathcal{C}$ will be a penalty for modifying the current policy and the threshold $\kappa$ will control the sparsity of the corresponding explanation).  Now, our goal is to maximize Equation~\ref{eqn:expected_return} while constraining the expected undiscounted costs: 
\begin{equation}
    \mathbb{E}_{\pi}\left[\sum_{t=0}^\infty \mathcal{C}(s_t,a_t)\right]\leq \kappa.
\end{equation}
For a discrete CMDP, the above problem can be written as a finite set of linear inequalities which can be solved using Linear Programming (LP)~\cite{altman1999constrained}.

The standard CMDP formulation may result in stochastic policies, which are harder to interpret and implement. Thus, we shall use the formulation of~\citet{dolgov2005stationary} to obtain deterministic policy solutions to the CMDP, which solves a a mixed integer LP (MILP) as follows:
\begin{equation}\label{eqn:cmdp_milp}
\scriptsize
\begin{array}{ll@{}ll}
\arg\max_x  &\displaystyle\sum\limits_{s,a}x(s,a)\mathcal{R}(s,a) \\
\text{subject to}& \displaystyle\sum\limits_{a}x(s',a)-\sum\limits_{a,s}x(s,a)\mathcal{T}(s'|s,a) = p_0(s'),
& \forall s'\in\mathcal{S} \setminus \mathcal{S}_T\\
& \displaystyle\sum\limits_{a,s}x(s,a)\mathcal{C}(s,a)\leq \kappa\\
& \displaystyle\sum\limits_{a}\triangle(s,a)\leq 1\\
& x(s,a)/M\leq \triangle(s,a),   \forall s\in\mathcal{S}, a\in\mathcal{A}\\
& x(s,a)\geq 0,   \forall s\in\mathcal{S}, a\in\mathcal{A}
\end{array}
\end{equation}
In the above,  $x$ is the dual variable that can be interpreted as the occupancy measure, which counts the expected discounted number of taking action $a$ in state $s$. $M$ is a constant to force $0\leq x(s,a)/M\leq 1$, which can be set to the maximum of the expected return in Equation~\ref{eqn:expected_return} with $\mathcal{R}=1$ (i.e. the maximal number of visits to a state). $\triangle_{s,a}\in\{0,1\}$ is a binary variable and the last three inequalities in Equation~\ref{eqn:cmdp_milp} ensure a unique selection of the action $a$.

The optimal policy can be calculated by:
\begin{equation*}
    \pi(s,a) = 
    \begin{cases}
    x(s,a)/\sum_{s}x(s,a) \quad \text{if }\sum_{s}x(s,a) >0\\
    \text{arbitrary}\quad \text{if }\sum_{s}x(s,a) =0
    \end{cases}
\end{equation*}

\section{Framework Part I: Explanation Generation}\label{sec:framework}
Our first contribution is a complete, global contrastive explanation generator for RL tasks given the current policy $\pi_b$ and the new policy $\pi_e$. In this section, we introduce each module of our explanation-generator with an illustrative toy example, shown in Figure~\ref{fig:toy_domain} (details in Appendix~\ref{apx_subsec:toy}).
Figure~\ref{fig:toy_mdp} depicts the toy MDP while~\ref{fig:pi_b} and~\ref{fig:pi_e}  highlight the actions of the current policy $\pi_b$ and the new policy $\pi_e$, respectively. We set $\pi_b$ and $\pi_e$ to be deterministic for pedagogical purposes.

In Section~\ref{subsec:state_identifier}, we first formalize how to compare two policies in terms of their behaviors. We define a \textit{diverging state}, at which we observe behavioral differences between $\pi_b$ and $\pi_e$. 
In Section~\ref{subsec:state_aggregator}, we abstract diverging states into human-interpretable \emph{regions} (sets of states) over which the behaviors of $\pi_b$ and $\pi_e$ differ.
Then, in Section~\ref{subsec:outcome_estimator}, we show how to compare policies $\pi_b$ and $\pi_e$ in terms of their expected outcomes.
Finally, in Section~\ref{subsec:explanation_generator}, we combine the comparison in terms of behaviors and expected outcomes to generate the global contrastive explanation in Form~\ref{explanation_form}.

\subsection{Diverging State Identification}\label{subsec:state_identifier}
 To compare $\pi_b$ and $\pi_e$ in terms of behaviors, we first identify states at which the two policies take different actions that lead to different next states. We call such states \emph{diverging states}. Because the policies and the environment can be stochastic, comparing $\pi_b$ and $\pi_e$ involves comparing corresponding distributions over actions and next states. In this work, we make a heuristic approximation of the stochastic policies and the environment by focusing on events that occur with high probability. While there are many ways to compare two distributions (e.g. hypothesis testing or divergence measures), we choose this heuristic because policies in real life are frequently nearly deterministic.

Specifically, we denote the most probable action proposed by policies $\pi_b$ and $\pi_e$, in state $s$, as $a_b$ and $a_e$, respectively:
\[a_b = \arg\max_{a\in\mathcal{A}} \pi_b(a|s),\ a_e = \arg\max_{a\in\mathcal{A}} \pi_e(a|s). \]
We denote the most probable next state under $\pi_b$ and $\pi_e$ as $s'_b$ and $s'_e$, respectively:
\[s'_b =  \arg\max_{s'\in \mathcal{S}} \mathcal{T}(s'|s,\pi_b),\ s'_e =  \arg\max_{s'\in \mathcal{S}} \mathcal{T}(s'|s,\pi_e). \]
Given any state $s$, we define a function $h_\text{id}:\mathcal{S}\xrightarrow{}\{0,1\}$ mapping a state $s$ to a label $1$ (i.e. $s$ \textit{is} a diverging state) if it satisfies \textit{both} of the following conditions:
\begin{equation}
\label{eqn:diverging_state}
    \begin{cases}
    a_b \neq a_e \ \text{or}\  |\pi_b(a_b|s) - \pi_e(a_e|s)| > \kappa_\pi\\
      s'_b \neq s'_e \ \text{or}\  |\mathcal{T}(s_b|s,\pi_b) - \mathcal{T}(s_e|s,\pi_e)| > \kappa_\mathcal{T}\\
\end{cases}    
\end{equation}
where $\kappa_\pi, \kappa_\mathcal{T}$ are domain-dependent constants.
The first condition states that the policies $\pi_b$ and $\pi_e$ are different when they take different actions in state $s$ with high probability. That is, either: (a) the most probable actions under $\pi_b$ and $\pi_e$ are different, or (b) the most probable actions are identical, but with very different associated probabilities. 
The second condition ensures that even if the policies take different actions with high probability, the agent should not end in the identical next state with high probability. For example, in the toy example, state $s_1$, $s_5$ are diverging states while $s_0$ is not a diverging state, as $s_0$ fails to satisfy the second condition (others states fail to satisfy the first one). 

\textbf{Application to Continuous Domains.}
In a large or continuous domain, computing $h_\text{id}(s)$ for a representative number of states can be computationally expensive.  Even if we use sampling, collecting enough samples to cover the state space $\mathcal{S}$ to approximate the function $h_\text{id}(s)$ may be prohibitive.

We get around this issue by computing $h_\text{id}$ only for states that are probable under our policies.  To do this, we generate trajectories using the recursive Algorithm~\ref{alg:collect_state}, in which whenever encountering a divergence state, we split the current trajectory into two by rolling out $\pi_b$ and $\pi_e$ separately. 
By rolling out trajectories switching between policies, the algorithm ignores states that cannot be reached by either policy. Note that we split the trajectory whenever we see policies are different, since we expect earlier differences in the trajectory to have more effect on the expected outcomes.

Algorithm~\ref{alg:collect_state} includes a hyper-parameter $d_\text{max}$ that controls the number of times we are allowed to split. This allows the user to trade-off between exploration of the state space at an early state and the computational complexity of the roll-out (which grows exponentially with the number of splits). In practice, we find that a small $d_{\max}$ often covers a sufficient amount of diverging states for later tasks in our framework. 

\begin{algorithm}
\caption{Diverging\_State\_Collection ($\pi_1$, $\pi_2$, $s$, $d$,  $d_\text{max}$)}
\label{alg:collect_state}
\begin{algorithmic}[1]
  \scriptsize
  \STATE Initialize $\mathcal{D}=(\mathcal{S},\mathcal{Y})$, $\mathcal{I}=\{\}$,$k=1$
  \IF{$d\ \geq d_{\max}$}
  \STATE return  $\mathcal{D}$
  \ENDIF
  \WHILE{$s \notin \mathcal{S}_T$}
  \STATE Observe the current state $s$
  \STATE $a_1 = \arg\max\pi_1(a|s),\ a_2 = \arg\max\pi_2(a|s)$
  \IF{$h_\text{id}(s,\pi_1, \pi_2) = 1$}
  \STATE $\mathcal{D}'$ = Diverging\_State\_Collection ($\pi_1$, $\pi_2$, $s$, $d+1$, $d_{\max}$)
  \STATE $\mathcal{D}'\xrightarrow{}\mathcal{D}$
  \IF{key $(a_1, a_2) \notin \mathcal{I}$} 
  \STATE add key: $\mathcal{I}[(a_1, a_2)] = k$, $k = k + 1$
  \ENDIF
  \STATE $(s,\ \mathcal{I}[(a_1, a_2)])\xrightarrow{}\mathcal{D}$
  \ELSE
  \STATE $(s,\ 0)\xrightarrow{}\mathcal{D}$
  \ENDIF
  \STATE Perform action $a_1$
  \ENDWHILE
\STATE return  $\mathcal{D}$
\end{algorithmic}
\end{algorithm}

In the batch setting, we identify the diverging states by calculating $h_\text{id}(s)$ for every state in the batch data using a similar algorithm (Appendix~\ref{apx_subsec:batch_collect_state}).

\subsection{Diverging State Aggregation}\label{subsec:state_aggregator}
After labeling the state space (Section~\ref{subsec:state_identifier}), we may have identified a large number of diverging states. 
As such, a complete enumeration of the diverging states is unlikely to be useful to end-users. 
Thus, to summarize the difference between diverging and non-diverging states, we aggregate the diverging states into \textit{diverging regions} and extract general descriptions of these regions.

To perform state aggregation, we train an interpretable, rule-based classifier to predict the behavioral difference between $\pi_b$ and $\pi_e$ given a diverging state $s$.  The split rules of this classifier correspond to our diverging regions.

Specifically, Algorithm~\ref{alg:collect_state} will return a dataset with $N$ samples $\{(s_n, y_n)\}_{n=1}^N$, where $s_n\in\mathcal{S}$ and labels $y_n\in \{0,\dots,K\}$. We set $y_n = 0$ if $s_n$ is not a diverging state and $y_n = k$ otherwise, with each $k\in\{1,\dots,K\}$ representing a unique action pair $(a_b, a_e)$. 
Let the state space $\mathcal{S}$ be partitioned into $\bigcup_{k=0}^K\mathcal{S}_k$ where $\mathcal{S}_k$ contains the samples with labels $y_n = k$. 
We learn an interpretable classifier $h_\text{aggr}: \mathcal{S}\xrightarrow[]{}\mathcal{Y}$ that takes the states $s_n$ as input and predicts the label $y_n$. From the classifier $h_\text{aggr}$, we extract rules for describing each partition $\mathcal{S}_k$.

We have many choices of interpretable classifiers to partition the space of diverging regions.  In this work, we adapt Boolean Decision Rules via Column Generation (BDCG)~\citep{dash2018boolean} to generate rule-based explanations for diverging regions (Appendix~\ref{apx_subset:bdcg}), as this is a simple interpretable model with desirable regularization properties. 
Other interpretable classifiers may involve developing self-explanatory models~\citep{quinlan1987simplifying,zhang2018interpretable,sabour2017dynamic} 
or provide other forms of global insights for pre-trained models~\citep{altmann2010permutation,karpathy2015visualizing}.  
We encourage the developers to use any interpretable classifier appropriate for their tasks.

\subsection{Expected outcome estimation}\label{subsec:outcome_estimator}
In the previous section, we describe how policies $\pi_b$ and $\pi_e$ differ in terms of actions. In this section, we quantify the difference between policies in terms of the expected outcomes for a set of user defined functions (Equation~\ref{eqn:expected_outcomes}).

In practice, we usually have enough data for a good estimate of the expected outcomes of $\pi_b$, $\hat{\mathbf{G}}_{\pi_b}$: we either have access to the environment $\mathcal{T}$ in the online setting or enough data collected from $\pi_b$ in the batch setting. But we may not have enough information to obtain a good estimate of the expected outcomes of $\pi_e$, $\hat{\mathbf{G}}_{\pi_e}$. In these situations, it is important to inform the end-users of our uncertainty over the estimated expected outcomes in addition to the estimates themselves. Thus, when comparing the expected outcomes $\hat{\mathbf{G}}_{\pi_b}$ and $\hat{\mathbf{G}}_{\pi_e}$, we incorporate the confidence interval $[l(\mathbf{G}_{\pi_e}),u(\mathbf{G}_{\pi_e})]$ into our estimate $\hat{\mathbf{G}}_{\pi_e}$.

In particular, given an initial state $s_0$, we compute the difference in expected outcomes $\mathbf{h}_\text{oc} = [h^{(0)}_\text{oc}),\dots,h^{(M-1)}_\text{oc})]^\intercal$ as:
\begin{equation*}
h^{(m)}_\text{oc}) = 
    \begin{cases}
    1 & \text{if } \hat{{G}}^{(m)}_{\pi_b} < l(G^{(m)}_{\pi_e})  (\text{$\pi_e$ is better})\\
    -1 & \text{if } \hat{G}^{(m)}_{\pi_b} > u(G^{(m)}_{\pi_b})  (\text{$\pi_b$ is better})\\
    0 & \text{otherwise (unknown difference)}
    \end{cases}
\end{equation*}

Given the transition dynamics $\mathcal{T}$, the expected outcomes estimates $\hat{\mathbf{G}}_{\pi_b}$ and the confidence intervals $[l(\mathbf{G}_{\pi_e}),u(\mathbf{G}_{\pi_e})]$  can be obtained using Monte Carlo roll-outs.
In the batch setting, those terms can be estimated using off policy evaluation  methods such as importance sampling~\cite{thomas2016data} or fitted-Q evaluation~\cite{hao2021bootstrapping}.
We choose model-based Bootstrapping~\citep{hanna2017bootstrapping}, since for our simple domains, we can assume enough data to learn a good transition dynamics estimate, $\hat{\mathcal{T}}$.
\begin{figure}
    \centering
    \includegraphics[scale=0.3]{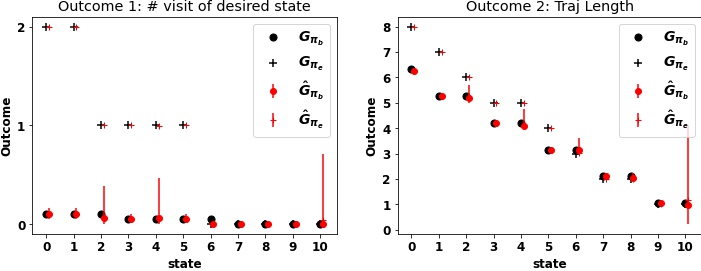}
    \caption{The true outcomes ${\mathbf{G}}_{\pi_b},\mathbf{G}_{\pi_e}$ and the estimated outcomes $\hat{\mathbf{G}}_{\pi_b}$, $\hat{\mathbf{G}}_{\pi_e}$ of discrete MDP: Black markers represent true outcomes while red ones represent estimates. The estimates are  accurate as black and red markers overlap.}
    \label{fig:toy_mdp_ope}
\end{figure}

For example, in the toy domain (Figure~\ref{fig:toy_domain}), we are interested in: (1) the number of times we visit the desired states (state $s_2,\ s_6$) and (2) the trajectory length. 
From Figure~\ref{fig:toy_mdp_ope}, we see that given the initial states $s_0 \cdots \ s_5$, policy $\pi_e$ visits the desired states more times (while yielding a longer trajectory) than policy $\pi_b$. But given the initial states $s_6\cdots s_{10}$, we are unsure of the differences in outcomes of the two policies. 

\subsection{Global Contrastive Explanation Generation}\label{subsec:explanation_generator}
Thus far, we have $h_\text{aggr}$ that aggregates the diverging states into human-interpretable regions, and $\mathbf{h}_\text{oc}$ that computes the differences between the expected outcomes of policies $\pi_b$ and $\pi_e$. We are now ready to generate a global contrastive explanation for $\pi_b$ and $\pi_e$ in the format of Form~\ref{explanation_form}, summarizing the behavioral and outcome differences between the two.

The detailed procedure is as follows (Algorithm~\ref{alg:generate_explanation}):
\begin{enumerate}
    \item Given an initial state $s_0$, we perform a roll-out following $\pi_e$. Here, we can use the true environment $\mathcal{T}$, or the estimate $\hat{\mathcal{T}}$. We track the set of diverging regions visited in our roll-out and define the trajectory of visited regions as the \textit{diverging path}: $\mathcal{P}(s_0)=\{s_{k_1},s_{k_2},\cdots\}$, where $s_{k_i}\in \mathcal{S}_k$ is defined in Section~\ref{subsec:state_aggregator}.
    \item We compute the difference in the expected outcomes of $\pi_b$ and $\pi_e$ using $h_\text{oc}$ from Section~\ref{subsec:outcome_estimator}.
    \item Different initial states may result in different diverging paths and outcome differences. To make the explanation more compact, we further aggregate the initial states by training an interpretable classifier (the adapted BDCG) that maps the initial state $s_0$ to the  tuple of the diverging path $\mathcal{P}(s_0)$ and outcome difference $h_\text{oc}(s_0)$.
    \item We extract the interpretable rules of the classifier from Step 3, which describes the relationship between an initial state and the resulting combination of the diverging path and outcome differences.
\end{enumerate}
\begin{algorithm}
\caption{Generate\_Explanations($h_\text{aggr}$, $\mathbf{h}_\text{oc}$, $\pi_b$, $\pi_e$)}
\label{alg:generate_explanation}
\begin{algorithmic}[1]
  \scriptsize
  \STATE Initialize $\mathcal{D}=(\mathcal{S},\mathcal{Y})$, $\mathcal{I}=\{\}$, $z=0$, episode length $L$
  \FOR{$s_0 \sim p_0$}
  \STATE Initialize $\mathcal{P}(s_0)=[]$
  \WHILE{$s\notin \mathcal{T}_T$}
  \STATE Observe the current state $s$
   \IF{$h_\text{aggr}(s)\neq 0$}
  \STATE $s\xrightarrow{}\mathcal{P}(s_0)$
  \ENDIF
  \STATE Perform $a\sim \pi_e$
  \ENDWHILE
  \IF{key $(\mathcal{P}(s_0), \mathbf{h}_\text{oc}(s_0)) \notin \mathcal{I}$} 
  \STATE add key: $\mathcal{I}[(\mathcal{P}(s_0), \mathbf{h}_\text{oc}(s_0))] = z$, $z = z + 1$
  \ENDIF
\STATE $(s_0, \mathcal{I}[(\mathcal{P}(s_0), \mathbf{h}_\text{oc}(s_0))])\xrightarrow{}\mathcal{D}$
\ENDFOR
\STATE Train a classifier $h_\text{exp}: \mathcal{S}\xrightarrow[]{}\mathcal{Y}$
\end{algorithmic}
\end{algorithm}

For the toy domain (Figure~\ref{fig:toy_domain}), we identify states $s_1,\ s_5$ as the diverging states (also as diverging regions because the MDP is discrete). We also compute the difference of the expected outcomes in Section~\ref{subsec:outcome_estimator}.
Together, this yields the following global contrastive explanation:

\begin{footnotesize}
\begin{Verbatim}[breaklines=true, breaksymbolleft=,commandchars=\\\{\},codes={\catcode`$=3\catcode`_=8}]
Starting from initial region $s_0$, in region $s_1$, doing action $1$ instead of action $0$ and then in region $s_5$, doing action 1 instead of action 0 will lead to longer trajectory but more visits to desired states;
	 
Starting from initial region $s_1$, in region $s_1$, doing action 1 instead of action 0 and then in region $s_5$, doing action 1 instead of action 0 will lead to longer trajectory but more visits to desired states;

Starting from initial region $s_2$, in region $s_5$, doing action 1 instead of action 0 will lead to longer trajectory and more number of visits to desired states;

Starting from initial region $s_3$, in region $s_5$, doing action 1 instead of action 0 will lead to longer trajectory and more number of visits to desired states;
	 
Starting from initial region $s_4$, in region $s_5$, doing action 1 instead of action 0 will lead to longer trajectory and more number of visits to desired states;
	 
Starting from initial region $s_5$, in region $s_5$, doing action 1 instead of action 0 will lead to longer trajectory and more number of visits to desired states;

Starting from initial region $s_6\cdots s_{10}$, two policies, $\pi_b$ and $\pi_e$ act the same.
\end{Verbatim}
\end{footnotesize}

\section{Framework Part II: Policy Optimization}
\label{sec:opt}
In Framework Part I, we describe a generator of complete, contrastive explanations in the RL setting. Now, we use these explanations to define interpretability constraints for policy optimization. That is, 
we illustrate how to modify the current policy $\pi_b$ such that (a) the improved policy $\pi_e$ increases expected returns (Equation~\ref{eqn:expected_return}) and (b) the contrastive explanation (as defined in Framework Part I) given $\pi_b$ and $\pi_e$ will be sparse. 
We formalize this goal as a CMDP that maximizes returns subject to a constraint on the number of changes made to the current policy $\pi_b$ (which corresponds to a sparse contrastive explanation).  
 
\begin{figure*}
     \centering
     \begin{subfigure}[b]{0.32\textwidth}
         \centering
         \includegraphics[width=\textwidth]{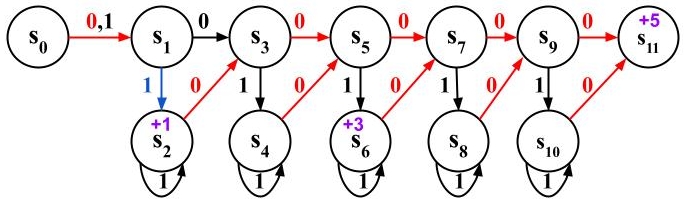}
         \caption{Optimal policy with constraint $\kappa=2$}
         \label{fig:cmdp_k1}
     \end{subfigure}
     \hfill
     \begin{subfigure}[b]{0.32\textwidth}
         \centering
         \includegraphics[width=\textwidth]{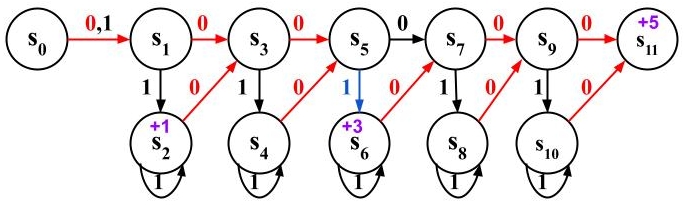}
         \caption{Optimal policy with constraint $\kappa=6$}
         \label{fig:cmdp_k2}
     \end{subfigure}
     \hfill
     \begin{subfigure}[b]{0.32\textwidth}
         \centering
         \includegraphics[width=\textwidth]{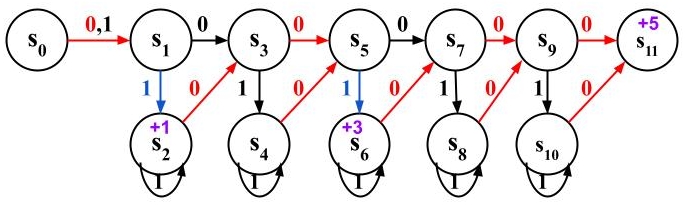}
         \caption{Optimal policy with constraint $\kappa=8$}
         \label{fig:cmdp_k3}
     \end{subfigure}
        \caption{Discrete MDP: Visualization of optimal deterministic policies when increasing $\kappa$. The purple number denotes the immediate rewards. Red arrows highlight the current policy behavior while the blue ones represent the changed behavior.
         The number of changes is aggregated over all initial states.}
        \label{fig:toy_cmdp}
\end{figure*}
 \begin{figure*}
     \centering
     \begin{subfigure}[b]{0.32\textwidth}
         \centering
         \includegraphics[width=\textwidth]{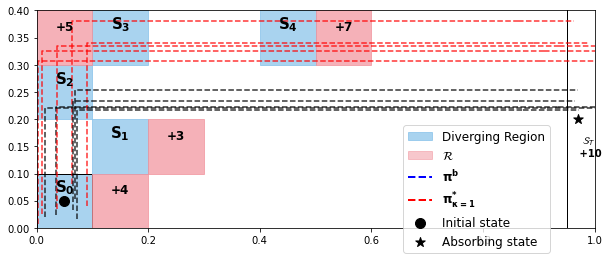}
         \caption{Optimal policy without constraint}
         \label{fig:2d_pib}
    \end{subfigure}
    \begin{subfigure}[b]{0.32\textwidth}
         \centering
         \includegraphics[width=\textwidth]{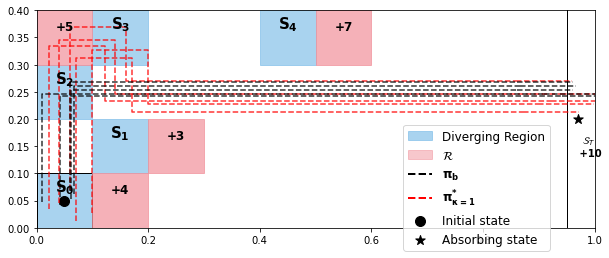}
         \caption{Optimal policy with constraint $\kappa=1$}
         \label{fig:2d_pie1}
    \end{subfigure}
    \begin{subfigure}[b]{0.32\textwidth}
         \centering
         \includegraphics[width=\textwidth]{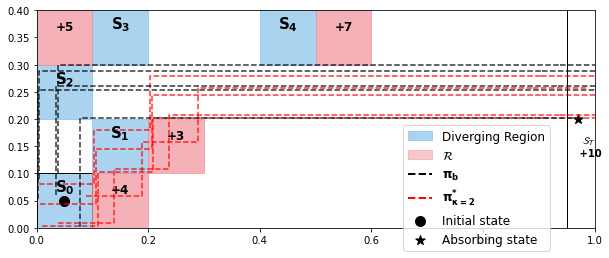}
         \caption{Optimal policy with constraint $\kappa=2$}
         \label{fig:2d_pie2}
    \end{subfigure}
    \caption{2D Navigation: Visualization of optimal policy trajectories when increasing $\kappa$. The rectangle with the black circle represents the initial state region while the one with the star represents the absorbing state region. Each blue box represents one diverging region while each red box represents regions with different rewards. Each step has a cost of $-0.001$.}
    \label{fig:2d}
\end{figure*}

\subsection{CMDPs for Sparse Changes in Discrete Domains}\label{subsec:cmdp_discrete}
In Section~\ref{sec:notation}, we introduced background information on CMDPs. Here, we recast our policy optimization problem over discrete domains as a CMDP. Since we want to encourage sparse modifications, we define the constraint as
$$\mathcal{C}(s, a) = \mathbbm{1}_{\{a\neq[\arg\max_{a}\pi_b(a|s)]\}},\quad \mathcal{C}(s_T,\cdot) = 0.$$
That is, every modification to the current policy $\pi_b$ occurred during a roll-out incurs a constant cost and we bound the expected number of changes to the current policy $\pi_b$. In addition, we assume the new policy is deterministic as it is more interpretable and executable.
We can control the length of the contrastive explanation by varying the threshold $\kappa$ (smaller $\kappa$ implies a more similar policy to $\pi_b$ and a sparser explanation, and vice versa).

\subsection{Extensions to Continuous Domains}
\label{subsec:cmdp_continous}
For continuous domains, we cannot rewrite the optimization using the LP formulation in Equation~\ref{eqn:cmdp_milp}, because now we have an infinite number of dual variables $x(s,a)$.

In this section, we show that, if we have a set of candidate improved policies (produced by RL algorithms or provided by experts), we can use this information to transform a continuous-state MDP into a discrete one to which we can apply the CMDP framework from Section~\ref{sec:notation}. Assuming a set of suggested policies is reasonable in our setting, as we can run RL algorithms to identify possible new policies that improves upon our current one. However, in this case, we still want to choose a new policy $\pi_{e}$ that maximizes expected returns with minimal changes to $\pi_b$.

To find the optimal $\pi_{e}$, we only need to study regions where candidate and current policies disagree, that is, the diverging regions defined in Section~\ref{subsec:state_aggregator}. Thus, we can recast the optimization problem as one over the discrete state space consisting of diverging regions.

Denote the set of candidate policies as 
$\{\pi_{e_j}\}_{j=1}^J$.  For each candidate policy $\pi_{e_j}$, we compare $\pi_{e_j}$ to the current policy $\pi_b$ and obtain a set of diverging regions following Section~\ref{subsec:state_identifier} and~\ref{subsec:state_aggregator}. We then collect all diverging regions, $\{\mathcal{S}_1, \mathcal{S}_2,\cdots, \mathcal{S}_K\}$, and denote the set containing the absorbing state region as well as the diverging regions as $\mathcal{S'} = \mathcal{S}_T\bigcup(\bigcup_{k=1}^K\mathcal{S}_k)$. We set $\mathcal{S'}$ as our new state space. 

By construction, our new state space $\mathcal{S'}$ is finite (ruling out severe pathology in $\pi_{e_j}$) and hence discrete. The action space remains the same. We define the transition dynamics $\mathcal{T}'(s_{k'}|s_k,a)$ on the new discrete state space as the probability of transitioning from the region $s_k$ to the region $s_{k'}$ by taking action $a$. We define a new reward function $\mathcal{R}'(s_k,a)$ by taking the average reward resulting from taking action $a$ in the region $s_k$. Details of the construction and estimation of the new discrete CMDP over diverging regions can be found in Appendix~\ref{apdx:discrete_cmdp}. Finally, we can recast the problem of finding an optimal policy $\pi_e$ as solving the  CMDP (Equation~\ref{eqn:cmdp_milp}) defined by our discretized MDP -- that is, we find $\pi_e$ that differs from the currently policy $\pi_b$ over a minimal set of diverging regions, and that attains the highest gain in expected return (averaged over these regions).

\section{Experiments \& Results}\label{sec:results}
Taken as a whole, our framework (explanation generation and policy optimization) improves the current policy with sparse and user-interpretable changes. We demonstrate our framework on a toy domain (the discrete MDP in Figure~\ref{fig:toy_domain}) and a continuous 2D navigation domain.

\subsection{2D Navigation Domain}
We present a 2D domain (Figure ~\ref{fig:2d}) where the agent starts from the initial region and aims to navigate to a goal region. The trajectories of the current policy $\pi_b$ are given in Figure~\ref{apx_fig:2d_pib}. We are also given two proposed policies $\pi_{e_1}$ (Figure~\ref{apx_fig:2d_pie1}) ,$\pi_{e_2}$ (Figure~\ref{apx_fig:2d_pie2}). As users, we want to stay in the desired region ($0.2\leq y <0.3$) while collecting as much reward as possible (Details of the domain in Section~\ref{apx_subsec:2d}).
\subsection{Explanation Generation} 
\label{subsec:results_exp}
We first show that our explanation-generator creates a compact explanation given two policies. For the discrete MDP, an example of the explanation is given in Section~\ref{subsec:explanation_generator}. For the 2D domain, we generate a contrastive explanation given the current policy $\pi_b$ and the optimal policy $\pi^*$ (Figure~\ref{fig:2d_pib}) without any sparsity constraints:
\vspace{-0.1cm}
\begin{flalign*}
\small
\begin{tabular}{@{}p{0.46\textwidth}}
``Starting from the initial region, going north instead of east when reaching region $0\leq x\leq 0.1, 0.2\leq y \leq 0.3$,
then going east instead of south when reaching region $0.1\leq x\leq 0.2, 0.3\leq y \leq 0.4$,
then going east instead of south when reaching region $0.4\leq x\leq 0.5, 0.3\leq y \leq 0.4$,
will result in less stay in the desired region ($0.2\leq y <0.3$) but more collected rewards.
\end{tabular}
\end{flalign*}
\subsection{Policy Optimization}
\begin{figure}[!htbp]
     \centering
     \begin{subfigure}[b]{0.23\textwidth}
         \centering
         \includegraphics[width=\textwidth]{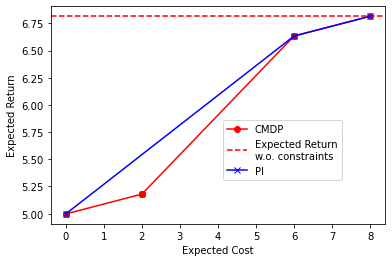}
         \caption{Discrete MDP}
         \label{fig:discrete_r}
     \end{subfigure}
          \begin{subfigure}[b]{0.23\textwidth}
         \centering
         \includegraphics[width=\textwidth]{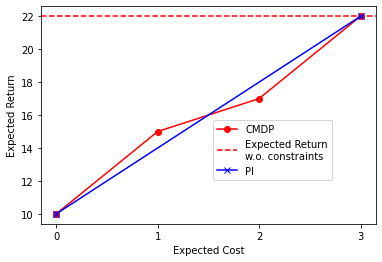}
         \caption{2D Navigation}
         \label{fig:2d_r}
     \end{subfigure}
        \caption{Policies found during optimization: $x$-axis denotes the expected cost of the policy while $y$-axis denotes the expected return. Red and blue lines represent our method (CMDP) and naive method (PI), respectively. In general, the expected return increases as the expected cost increases.
         We see that PI can only find a subset of intermediate policies while our method can find all.}
        \label{fig:expected_return}
\end{figure}

We compare our sparse-explanation-constrained optimization procedure in Section~\ref{sec:opt} to a naive method, Policy Iteration (PI), in which we iteratively evaluate and improve the current policy until convergence. During policy improvement, we greedily choose the action with the largest gain. 

In Figure \ref{fig:expected_return}, we see that our optimization procedure is able to find all intermediate policies that improve upon the current policy (in terms of reward) as well as the optimal solution (Figure~\ref{fig:toy_cmdp},~\ref{fig:2d}).
Comparing to us, PI can only find some of the intermediate solutions. For 2D domain, PI can only find the optimal solution. 
Although the optimal policy collects the largest reward, the corresponding contrastive explanation is long (Section~\ref{subsec:results_exp}) even for a simple domain like our 2D navigation.  Our method allows users to select the policy based on their own preference for the interpretability of the contrastive explanation as well as their considerations of additional outcomes. For example, for the 2D Navigation, the users may prefer the policy when $\kappa = 2$: although it collects less rewards, it stays longer in the desired region, $0.2\leq y <0.3$ (Figure \ref{fig:2d_pie2}).

\section{Discussion and Conclusion}
In this work, we build a framework that improves a current policy (with which the user is already familiar) such that the contrastive explanation given the current and improved policy will be sparse and human interpretable. Our contribution is two-fold: (1) we generate complete, user-friendly global contrastive explanations for two RL policies and (2) we optimize the current policy while explicitly constraining the length of the resulting contrastive explanation.

Compared to existing methods for generating contrastive explanations, our explanations are more complete while still being compact.
For example, in~\citeauthor{van2018contrastive}, the authors study the difference in the paths generated by rolling out two policies and only explain short-term outcomes. On our discrete MDP domain, their method would generate the following explanation:

\begin{flalign*}
\small
\begin{tabular}{@{}p{0.42\textwidth}}
``Starting from the initial state $s_0$, $\pi_e$ will reach state $s_1$ and perform action $1$, which visits the desired state $s_2$ but increases the trajectory length by $1$. Then, $\pi_e$ will reach state $s_5$ and perform action $1$, which visits the desired state $s_6$ but increases the trajectory length by $1$."
\end{tabular}
\end{flalign*}
In~\citeauthor{sukkerd2020tradeoff}, the authors explain the outcome difference without telling users why the difference occurs:
\begin{flalign*}
\small
\begin{tabular}{@{}p{0.42\textwidth}}
``Starting from the initial state $s_0$, we can visit the desired states $(s_2,s_6)$ more times by carrying out the alternate policy $\pi_e$. However, this would make the trajectory longer."
\end{tabular}
\end{flalign*}
In contrast, our contrastive explanation is more compact and complete (see Section \ref{subsec:explanation_generator}) as we explain the global behavioral differences of two policies and the resulting differences in their long term outcomes.

Our sparse-explanation-constrained policy optimization procedure allows the end-users to explicitly control the trade-off between improved returns, interpretability and other desirable outcomes. While sparse optimization has appeared across counterfactual explanation literature, we are, to our knowledge, first to consider sequential decision-making tasks under sparsity constraints applied to explanations.

In future works, we plan to apply our framework to large healthcare domains, with the potential of helping clinicians understand new treatment plans and potentially improving current plans.
\section{Acknowledgement}
JY and FDV acknowledge the support from NSF IIS-2007076.

\bibliography{example_paper}
\bibliographystyle{icml2022}

\newpage
\appendix
\onecolumn
\providecommand{\upGamma}{\Gamma}
\providecommand{\uppi}{\pi}
\section{Experiment Domain}
\subsection{Toy Discrete MDP}\label{apx_subsec:toy}
\begin{figure*}[h]
    \centering
    \includegraphics[scale=0.25]{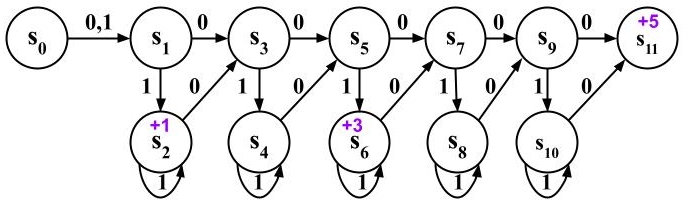}
    \caption{Discrete MDP Definition}
    \label{apdx_fig:discrete_mdp}
\end{figure*}
The discrete MDP has $11$ states and binary actions ($|\mathcal{S}|=11,\ \mathcal{A}\in\{0,1\}$).
$s_{11}$ is the absorbing state with the initial state uniformly sampled from other states ($p_0(s)= \text{Cat}(\frac{1}{10}\vec{1})$). 
The transition dynamics $\mathcal{T}$ is defined as in Figure~\ref{apdx_fig:discrete_mdp}.

During explanation generation, users are interested in (a) the trajectory length (2) the number of visits to the desired states ($s_2, s_6$). During optimization, we aim to optimize the expected return with the reward function defined as 
\begin{equation*}
\mathcal{R}(s,a) = 
    \begin{cases}
    1 \quad \text{if } s=s_1, a=1\\
    3 \quad \text{if } s=s_5, a=1\\
    5 \quad \text{if } s\in \{s_9,s_{10}\}, a=0\\
    -0.001\quad \text{o.w. }
    \end{cases}
\end{equation*}

\subsection{2D Navigation}\label{apx_subsec:2d}
 \begin{figure*}[h]
     \centering
     \begin{subfigure}[b]{0.32\textwidth}
         \centering
         \includegraphics[width=\textwidth]{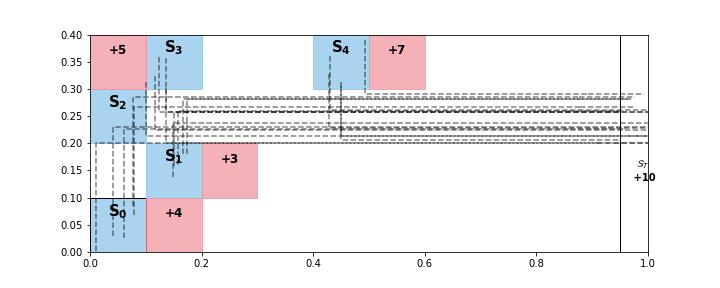}
         \caption{Current Policy $\pi_b$}
         \label{apx_fig:2d_pib}
    \end{subfigure}
    \begin{subfigure}[b]{0.32\textwidth}
         \centering
         \includegraphics[width=\textwidth]{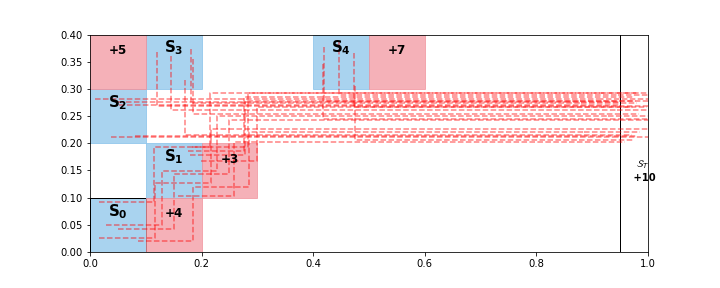}
         \caption{Proposed Policy $\pi_{e_1}$}
         \label{apx_fig:2d_pie1}
    \end{subfigure}
    \begin{subfigure}[b]{0.32\textwidth}
         \centering
         \includegraphics[width=\textwidth]{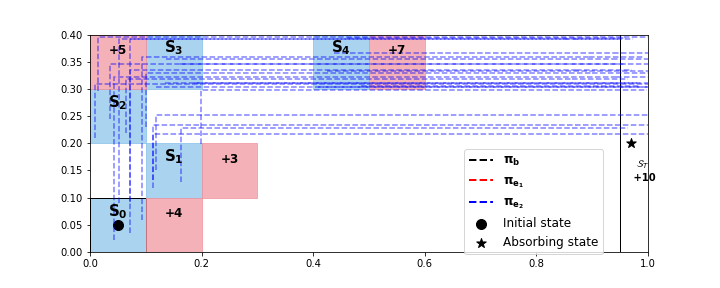}
         \caption{Proposed Policy $\pi_{e_2}$}
         \label{apx_fig:2d_pie2}
    \end{subfigure}
    \caption{2D Navigation Domain: Visualization of behaviors of current policy $\pi_b$ and two proposed policies $\pi_{e_1}$,$\pi_{e_2}$. The blue boxes represent diverging regions with red boxes represent regions with positive rewards. $\pi_b$ and $\pi_{e_1}$ differ in region $\mathcal{S}_0$ and $\mathcal{S}_2$. $\pi_b$ and $\pi_{e_2}$ differ in region $\mathcal{S}_1$ and $\mathcal{S}_3$.}
    \label{apdx_fig:2d}
\end{figure*}
In the continuous 2D Navigation domain, the state space is defined by the coordinates $s=(x,y)$. The agent can either go east ($a=E$), north ($a=N$) or  south ($a=S$). The initial state is uniformly sampled ($x\sim\text{Unif}(0,0.1),y\sim\text{Unif}(0,0.1)$) and the agent aims to navigate to the goal region ($x>0.95$).
The transition dynamics is defined as
\begin{equation*}
    \begin{cases}
    x' = x+0.1 \quad\text{if } a=E\\
    y' = y+0.1 \quad\text{if } a=N\\
    y' = y-0.1 \quad\text{if } a=S\\
    \end{cases}
\end{equation*}
The reward function is defined as 
\begin{equation*}
\mathcal{R} = 
    \begin{cases}
    4 \quad \text{if } x\in[0.1,0.2] \text{ and }y\in[0,0.1]\\
    3 \quad \text{if } x\in[0.2,0.3] \text{ and }y\in[0.1,0.2]\\
    5 \quad \text{if } x\in[0,0.1] \text{ and }y\in[0.3,0.4]\\
    7 \quad \text{if } x\in[0.5,0.6]\text{ and }y\in[0.3,0.4]\\
     10 \quad \text{if } x>0.95\\
    -0.001\quad \text{o.w. }
    \end{cases}
\end{equation*}

The current policy $\pi_b$ and two proposed policies $\pi_{e_1},\pi_{e_2}$ (Figure~\ref{apdx_fig:2d}) is defined as
\begin{multicols}{3}
\footnotesize
\noindent    
\begin{equation*}
\pi_b(a|s)=
    \begin{cases}
    S \quad&\text{if } x \in [0.1,0.2]\cup[0.5,0.6]\\
    &\text{and } y > 0.3\\
    N \quad&\text{else if } y < 0.2\\
    E \quad&\text{o.w. }\\
    \end{cases}
\end{equation*}
\noindent    
\begin{equation*}
\pi_{e_1}(a|s)=
    \begin{cases}
    E \ &\text{if } x\in[0,0.1]\\
    &\text{and }y\in[0.,0.1] \\
    E \ &\text{else if } x\in[0.1,0.2]\\
    &\text{and }y\in[0.1,0.2] \\
    S \quad&\text{else if } x \in [0.1,0.2]\cup[0.5,0.6]\\
    &\text{and } y > 0.3\\
    N \ &\text{if } y < 0.2\\
    E \ &\text{o.w. }\\
    \end{cases}
\end{equation*}
\noindent    
\begin{equation*}
\pi_{e_2}(a|s)=
    \begin{cases}
    N \ &\text{if } x\in[0,0.1]\text{ and }y \in[0.2,0.3]\\
    N&\text{else if } y<0.2\\
    E &\text{o.w. }\\
    \end{cases}
\end{equation*}
\end{multicols}
\section{Algorithms}
\subsection{Collect Diverging States (Batch)}\label{apx_subsec:batch_collect_state}
Suppose we are given a batch of data $\mathcal{D}_b$, we collect the diverging state using the following algorithm
\begin{algorithm}
\caption{Batch\_Diverging\_State\_Collection ($\pi_1$, $\pi_2$, $\mathcal{D}_b$)}
\label{alg:batch_collect_state}
\begin{algorithmic}[1]
  \scriptsize
  \STATE Initialize $\mathcal{D}$,  $\mathcal{I}=\{\}$,$k=1$
  \FOR{$s\in \mathcal{D}_b$}
  \STATE $a = \pi_1(a|s),\ a_2 = \pi_2(a|s)$
  \IF{$f(s,\pi_1, \pi_2) = 1$}
  \IF{key $(a_1, a_2) \notin \mathcal{I}$} 
  \STATE add key: $\mathcal{I}[(a_1, a_2)] = k$, $k = k + 1$
  \ENDIF
    \STATE $(s,\mathcal{I}[(a_1, a_2)])\xrightarrow{}\mathcal{D}$ 
  \ELSE
    \STATE $(s,0)\xrightarrow{}\mathcal{D}$ 
  \ENDIF
  \ENDFOR
\STATE return  $\mathcal{X}$, $\mathcal{Y}$
\end{algorithmic}
\end{algorithm}
\subsection{Adapted Boolean Decision Rules via Column Generation (BDCG)}\label{apx_subset:bdcg}
BDCG is a binary interpretable classifier that learns Boolean rules in either disjunctive normal
form (DNF, OR-of-ANDs) or conjunctive normal
form (CNF, AND-of-ORs). We adapt BDCG so that it can be used for our multi-classification tasks. 

Suppose we are given a data set consisting of $N$ samples $(\mathbf{x}_n, y_n)$ with labels $y_n = 1,\cdots,K$. Let the set $\{1,\cdots, N\}$ be partitioned into $\bigcup_{k=1}^K {\mathcal{Z}_k}$ where $\mathcal{Z}_k$ contains the indices of samples with the label $y_n=k$. Without loss of generality, we sort $\mathcal{Z}_k$ by the set size and assume that $|\mathcal{Z}_1|\leq |\mathcal{Z}_2|\leq \cdots \leq |\mathcal{Z}_k|$. The pseudo-code of the adapted BDCG is as follows
\begin{algorithm}
\caption{BDCG\_for\_Multi-classification ($\mathcal{X},\mathcal{Y},\mathcal{Z}$)}
\label{alg:batch_collect_state}
\begin{algorithmic}[1]
  \scriptsize
  \STATE Initialize $\mathcal{H}=\{\}$
  \FOR{$k=1,\cdots,k-1$}
  \STATE Recreate labels $\mathcal{Y}'$ ($y_n=0$ if $n\in\mathcal{Z}_k$ and $y_n=1$ otherwise)
 \STATE $h_k=$BDCG($\mathcal{X},\mathcal{Y'}$)
 \STATE $h_k \xrightarrow[]{}\mathcal{H}$
  \STATE $\mathcal{X} = \mathcal{X}\backslash\{x_n:n\in\mathcal{Z}_k\}$,$\mathcal{Y} = \mathcal{Y}\backslash\{y_n:n\in\mathcal{Z}_k\}$
  \ENDFOR
\STATE return  $\mathcal{H}$
\end{algorithmic}
\end{algorithm}

\section{Construct a discrete MDP for continuous domains}\label{apdx:discrete_cmdp}
Suppose we have collected $K$ diverging regions and denote them as 
$\{\mathcal{S}_1, \mathcal{S}_2,\cdots, \mathcal{S}_K\}$. We construct a discrete MDP where the state space is defined as $(\bigcup_{k=1}^K \mathcal{S}_k)\bigcup\mathcal{S}_T$ ($\mathcal{S}_T$ is the absorbing state region in the original MDP). The action space remains the same as the original MDP.
Recall that given an initial state $s_0$, a diverging path is defined as the set of diverging regions visited in a roll-out and is denoted as,
$\mathcal{P}(s_0)=\{s_{k_1},s_{k_2},\cdots, s_T\}$.

Given a set of trajectories generated by following the current policy $\pi_b$ and the proposed policies $\{\pi_{e_j}\}_{j=1}^J$, we can estimate the new transition dynamics as
\begin{equation*}
    \mathcal{T}(s_{k'}|s_k, a) = \frac{\sum\mathbbm{1}_{\{s_{k_t}=s_k,a_t = a,s_{k_{t+1}\}}=s_{k'}\}}}{\sum\mathbbm{1}_{\{s_{k_t}=s_k,a_t = a\}}}
\end{equation*}
and the new reward function as 
\begin{equation*}
    \mathcal{R}(s_k,a) = \frac{\sum\mathcal{R}_{s_k:,a}}{\sum\mathbbm{1}_\{{s_{k_t}=s_k,a_t = a}\}}
\end{equation*}
where $\sum\mathcal{R}_{s_k:,a}$ denotes the total of the sequence of rewards collected starting from the diverging region $s_k$ until reaching the next  diverging region or the absorbing region.

\end{document}